# GAN Computers Generate Arts? A Survey on Visual Arts, Music, and Literary Text Generation using Generative Adversarial Network

Sakib Shahriar
Department of Computer Science and Engineering
American University of Sharjah, UAE
b00058710@aus.edu

*Abstract—* **"*Art is the lie that enables us to realize the truth.*" – Pablo Picasso. For centuries, humans have dedicated themselves to producing arts to convey their imagination. The advancement in technology and deep learning in particular, has caught the attention of many researchers trying to investigate whether art generation is possible by computers and algorithms. Using generative adversarial networks (GANs), applications such as synthesizing photorealistic human faces and creating captions automatically from images were realized. This survey takes a comprehensive look at the recent works using GANs for generating visual arts, music, and literary text. A performance comparison and description of the various GAN architecture are also presented. Finally, some of the key challenges in art generation using GANs are highlighted along with recommendations for future work.**

## I. Introduction

Art, a manifestation of human creativity, has always been an essential component of human culture. The artistic expression of man has allowed us to study our history and realize our progress through time. For centuries, humans have focused on creating art to express their imagination, thoughts, memories, and ideas. The art form is usually classified into visual arts and performing arts. Visual art mainly contains works involving painting, sculpture, and architecture [1]. Performing arts include theatre, film, music, dance, and literature. Various definitions of art have been proposed throughout history such as imitation, representation, medium for the transmission of feelings, and intuitive expression [2]. However, for the purpose of this work, the focus will be on computer-generated arts, often referred to as generative art and computer art. It is generative because the art was at least in some part produced automatically by a computer program [3]. According to the American artist Sol Lewitt, 'The idea becomes a machine that makes the art' [4].

As computers increased in their capabilities from simple calculations to gaming applications, a lot of speculations regarding the creative ability of computers were raised. Could computers create a painting or write poems? In 1963, the Computers and Automation magazine launched a computer art contest with computer-generated arts being featured on the magazine's cover [5]. Figure 1 displays the 'hummingbird' composition - winner of the 1968 contest.

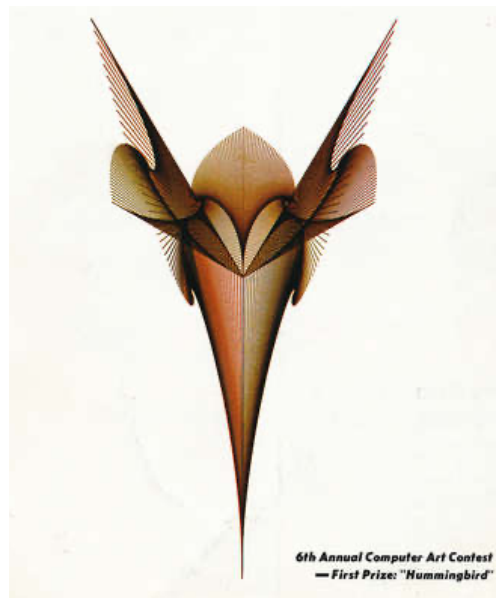

Figure 1 Hummingbird – Winner of 1968 Computer-generated Arts Competition [6]

Despite the early promising developments, computer-generated arts stagnated over the years. These artworks were very uniform and lifeless, unlike human arts. The major drawbacks included the use of programming languages and extensive human supervision. In this context, the computer's role was to suggest syntheses and leave the human expert to complete the feedback loop by making the final decision of accepting or rejecting the suggestion [7]. Moreover, there was not much scope for computers to generate literary works such as poetry. Music generation was limited and followed a similar approach to that of visual arts. Functions for amplitude, frequency, and duration of a sequence of notes were drawn on a cathode-ray tube with a light pen and the computer provided generations by combining these functions using simple algorithms [7]. Therefore, as long as computers continued to learn like computers and not humans, they would be constrained on their creative abilities. The challenging aspect of computer art generation comes from the fact that human supervision is required at some level. If this is the case, no matter how powerful computers get, they cannot independently produce artworks. The solution can only be achieved if computers are trained like humans, i.e., to learn from experience, using lots of data and without explicit programming. This is where deep learning and more specifically generative adversarial networks (GANs) can be effective.

In machine learning, the objective is to provide a given system with the ability to learn from experience or data without requiring explicit programming [8]. Deep learning is considered a subset of machine learning that primarily makes use of variations of artificial neural networks (ANN). Organized in a layered hierarchy of concepts (the term 'deep' coming from the depth of the layers), complex concepts are defined in terms of simpler ones and more abstract representations are gathered using less abstract ones [9].

Generative Adversarial Networks (GANs) use deep learning architectures to facilitate generative modeling. In this approach, the goal of the model is to generate new examples of data that would not be distinguishable by humans as data coming from the real set. This is achieved upon successful training where the adversarial network can identify patterns in the data and learn the distribution of the dataset. GANs primarily consist of two deep learning models, namely the generator and the discriminator. The generator is trained to generate new data points from given noise and the discriminator identifies the data points to be either real or fake. Both the models are trained together in a zero-sum game with the end goal of the discriminator being fooled by the generator about half the time [10]. GANs have already achieved remarkable results in various applications including the generation of photorealistic images, scenes, and people which are non-recognizable as fakes even to humans.

The idea of computers generating art without human supervision can be realized using GANs, unlike previous approaches. The fact that GANs can automatically learn to generate new examples from a given set of data proves to be a very useful component in computer art generation. Also, due to the advancement of technology and digitization, the two main requirements of GANs, datasets and computing power, have become widely available. As explained in the previous section, the older approaches to computer art generation relied on extensive human supervision and programming that led to the production of limited and uniform arts. Therefore, this makes GANs, a natural candidate to solve the difficult problem of computer-generated arts.

Existing works have also utilized genetic programming for art generation [11], which is based on the concept of evolution. Such approaches iteratively generate examples and then use a fitness function to evaluate them. This is followed by further modifications to improve the performance in the next iteration. As the authors in [11] points out, the main challenge using this approach is the question of 'how to write a logical fitness function that has an aesthetic sense'. Therefore, it is not suitable to rely on a programming approach to art generation as this will restrict the overall creativity and diversity of the generations. GANs, on the other hand, are not restricted to any such conditions, and therefore has the potential to create more realistic arts.

Although there exist several survey papers related to the proposed work, their scope is either too broad or too narrow. For instance, [12] provides a comparison of deep learning tools in the context of visual arts. The focus of the survey was on classification, evaluation, and generation of visual arts using many deep learning and artificial intelligence techniques. Conversely, [13] provides a review of free-hand sketch using deep learning tools. While a very specific review maybe useful for few researchers, in this survey we increase the scope by including three main art forms. At the same time, we keep the paper focused by reviewing art works generated using a specific deep learning model, i.e., GANs. We hope this work will inspire researchers and artists to collaborate on advancing the field of computer-generated arts using GANs. Following are the key contributions of this paper:

- It provides an overview of the primary GAN architectures for generative arts.
- It provides a comparison of the recent works in generative arts categorized by visual arts, music, and literary text generation along with their impacts.
- It proposes a discussion about the limitations of the existing works and provides future research directions.

## II. BACKGROUND

In this Section, the necessary background information including relevant GAN architectures will be discussed. Furthermore, common loss functions will be defined.

### A. Generative Adversarial Netowrks

GANs, first introduced by [14], can generate new content based on a min-max game between two networks, the generator and the discriminator. The generator tries to generate fake samples, having similar distribution to the examples in the real dataset and continues to improve its network to trick the discriminator. Meanwhile, the discriminator tries to distinguish between the real and fake samples. Figure 2 depicts a basic GAN architecture consisting of a single generator and discriminator. The generator has no access to data points from the real dataset. It minimizes its error by receiving feedback from the discriminator. Meanwhile, the discriminator has access to the ground truths, whether the data came from the generator or real dataset, which it can use to minimize its error.

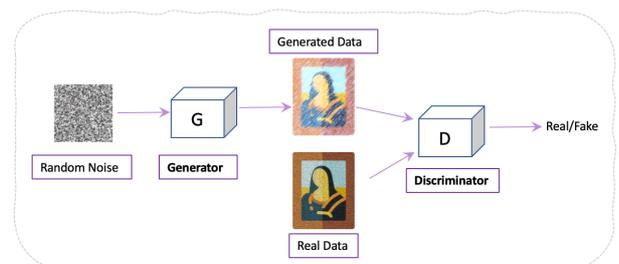

Figure 2 A Basic GAN Architecture

More sophisticated GAN architectures can make use of labels to generate data points for a specific category. This is useful because in the standard GAN, data points are generated only based on the input noise. However, unconditional generation of data points is not suitable for many applications. Next, various GAN architectures relevant to art generation is discussed.

*1) Conditional GAN:* Extended from the regular GAN, it is a conditional architecture if the generator and discriminator are conditioned on auxiliary information such as class labels [15]. The auxiliary information is combined in a joint representation with the noise in the basic GAN, allowing the generator to control the generation of data points based on the input condition. This type of architecture is suitable for multimodal generation of data points. For instance, conditional GANs can be used to generate various genres of music including jazz, rock, and classic.

*2) Deep Convolutional GAN:* Commonly known as DCGAN [16], the generator and discriminator in this architecture is made up of convolutional networks. Strided and fractionally strided convolutions are used to learn the spatial up sampling and down sampling operators. This facilitates the mapping from image space to lower dimensional latent space, and from image space to a discriminator [17]. Given the success of convolutional neural network in image and video classification in recent years, DCGAN remains a suitable architecture for image generation applications.

*3) Recurrent Adversarial Netwokrs:* In this architecture, a recurrent computation is obtained by unrolling the gradient descent optimization [18]. The two main components of this architecture are the convolutional encoder which extracts images of the current "canvas" and the decoder which decides whether or not to update the "canvas" by looking at the code for the reference image. Recurrent architectures are suitable for sequential and time-dependent data and is generally used for text and audio related applications.

Other common GAN architectures include InfoGAN [19], Laplacian GAN [20], and Wasserstein GAN [21]. For a comprehensive comparison of GAN architectures, the readers are encouraged to refer to [17] and [22].

*B. Common Loss Functions*

GAN training in a broader sense is based on a min-max game between two networks, the generator, and the discriminator. More precisely, each of the two networks have their losses or objective functions defined which it tries to iteratively optimize. The generator tries to adjust its weight parameters for every epoch or iteration that would ultimately lead it to generate realistic data points. The discriminator meanwhile attempts to tune its weight parameters such that it is capable of distinguishing between the real and generated samples. The loss function is what determines how well the models can get to the optimal set of parameters. The most common loss function is known as the cross-entropy loss, and it is defined in Equation 1:

$$H(p,q) = -\sum_{\forall x} p(x) \log(q(x)) \quad (1)$$

where $p(x)$ is the true distribution and $q(x)$ is the estimated distribution. Binary cross-entropy loss is a variation of cross-entropy loss, specifically used when dealing with two categories. It is commonly used in discriminators and can be defined as follows in Equation 2:

$$-\frac{1}{N} \sum_{i=1}^{N} y_i \cdot \log(p(y_i)) + (1 - y_i) \cdot \log(1 - p(y_i)) \quad (2)$$

where $p(y_i)$ is the probability of belonging to the real class and $1 - p(y_i)$ is the probability of belonging to the fake class. The other popular loss is the mean squared error (MSE) loss. This is defined in Equation 3:

$$\text{MSE} = \frac{1}{N} \sum_{i=1}^{N} (y_i - \hat{y}_i)^2 \quad (3)$$

where $y_i$ is the actual value and $\hat{y}_i$ is the predicted value for $N$ training examples. This loss function is typically used in the least square GAN (LSGAN). KL divergence loss is also widely used, and it can be defined in Equation 4:

$$\text{KL}(P \parallel Q) = \sum_{x \in \mathcal{X}} P(x) \log \left( \frac{P(x)}{Q(x)} \right) \quad (4)$$

Where $P(x)$ and $Q(x)$ are the two distributions being compared. The intuition behind KL divergence is to determine how much a probability distribution differs from another one. More recently, Wasserstein distance has gained interest as a loss function, especially in GAN training. It is defined in Equation 5:

$$l_1(u,v) = \inf_{\pi \in \Gamma(u,v)} \int_{\mathbb{R} \times \mathbb{R}} |x - y| \mathrm{d}\pi(x,y) \quad (5)$$

Where $u$ and $v$ are the distributions whose distance we are computing. $\Gamma(u,v)$ is the set of distributions in $\mathbb{R} \times \mathbb{R}$ whose marginals are u and v on the first and second factors respectively [23].

III. RECENT ADVANCES IN ARTS GENERATION USING GANs

In this section we provide a discussion and comparison of the recent works using GANs for generating various art forms including visual arts, music, and literary texts.

*A. Organization*

In this survey, the main organization will be based on the generation of three types of artworks. These are visual arts generation, music generation, and literary text generation. Within each of these sub-applications, the paper will dive deep into the main research challenges and provide a comparison of results. Furthermore, the loss function used, the GAN type, and architectural details will also be highlighted. Figure 3 provides a graphical illustration of the framework for organization of this survey paper.

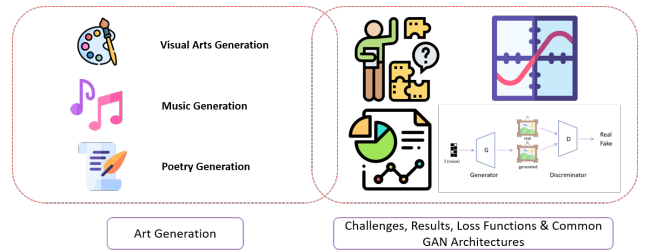

Figure 3 Framework of the Survey

*B. Visual Arts Generation using GANs*

This section presents a comparison of GAN approaches for visual arts generation. Lie *et al*. [24] presented a conditional GAN framework that can automatically generate painted cartoon images from a given sketch. The proposed architecture is a supervised learning approach such that given a black-and-white sketch, the model can create a painted colorful image. The training dataset contains the pair of sketches as well as ground truth colored images. To avoid the information loss that can occur using a pure encoder-decoder network, this work employed a U-net structure for the generator which concatenates the layers of the encoder to the corresponding decoder layer. The discriminator on the other hand solely consists of encoder units to classify the input sketch-image pair as 'real' or 'fake'. The authors utilized two datasets, namely the *Minions* dataset containing 1100 colored minions and the *Japanimation* dataset containing 6000 pictures of Japanese anime with 90% of the data being used for training and the remaining 10% for evaluation. For evaluation, 55 volunteers were asked to pick their most and least favorite generations from four different models. Based on this, the proposed architecture outperformed the three existing ones. A similar conditional GAN architecture using a U-Net generator for generating shoe image from a given shoe sketch is presented in [25]. However, this paper failed to provide appropriate evaluation for the generations and the samples generated by the models were not of high visual quality. Similarly, the authors in [26] implemented a GAN-based solution for generating synthetic images that are fully colored and textured from a given sketch. However, unlike previous works, their architecture allows the user to specify a style which could be based on the artist's name or a style category. The generator utilizes a U-Net architecture whereas the discriminator produces predictions of a style vector, sketch, and 'real' or 'fake' indicator by using the input image sketch and the image produced by the generator. The dataset used contains 10k images of 55 different styles such as impressionism, realism, and symbolism. Fréchet Inception Distance (FID) is used to evaluate the generations and classification accuracy is used to evaluate the style classification. In terms of both these metrics, the proposed architecture outperformed the existing works. For human evaluation, 100 participants were asked to evaluate between the proposed model and the baselines, with 63.3% preferring the generations made by the proposed model. Figure 4 illustrates some of the generations made using the proposed model based on the input sketch and the style including the name of the artist and the style type.

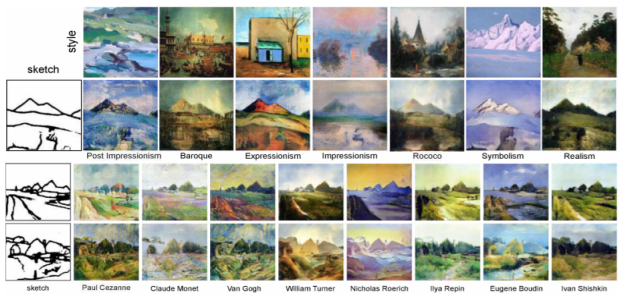

Figure 4 Sample Generations from the Proposed Sketch to Image Model in [26]

Nakano [27] introduced a framework for generating paintings based on learning brushstrokes. First, a variational autoencoder was used to learn the latent space of brushstrokes. However, the brushstrokes generated by the decoder contained a 'smudging' effect and the brushstrokes were smoothed out, which is not very realistic. This led the author to experiment with GANs for generating more realistic brushstrokes. Instead of giving the generator random noise as input as it is in typical GAN, the action space was provided in this approach. The action space is the set of parameters for controlling the painting environment, including brush size and color, to train the generator. The brushstrokes generated by the GAN architecture were rougher and more realistic. However, the author did not elaborate on the generator and discriminator architectural parameters. More importantly, the paper failed to provide any evaluation despite presenting some realistic paintings and generated samples.

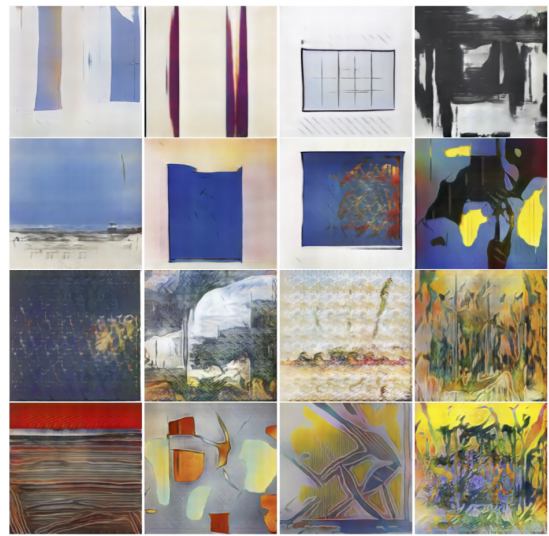

Figure 5 Samples from Proposed Architecture in [28]

In [28], the authors proposed a modified version of DCGAN for generating arts without any condition. The discriminator is responsible for classifying whether the generated example is an art or not and also classifying the art style. The generator besides taking as input the random vectors, also receives feedback from the discriminator. This includes whether or not the generation was an art and the style ambiguity, which indicates how well the discriminator can classify the generated art into established styles. The authors utilized the *WikiArt* dataset which consists of more than 80k paintings from 1119 artists ranging from the 15[th] to 20[th] century. For qualitative evaluation, human participants were asked to answer if they think the generations are created by an artist or a computer. They were also asked to rate it from 1 to 5 based on how much they liked a specific image. When compared to the baseline DCGAN, the proposed work had better performance with 53% respondents believing that the works were generated by an artist as opposed to 35% for DCGAN. More interestingly, it achieved an average of 3.2/5 score which is higher than those generated by actual artists (3.1/5). Further case studies were conducted to evaluate the originality and complexity of the generations. Figure 5 illustrates some of the sample arts from this work.

Tian *et al.* [29] introduced a pre-modern Japanese art facial expression dataset. This dataset contains more than 5500 RGB images alongside labeled gender and social status. Social status includes noble and warrior and can be used for supervised learning applications. The authors then explored with an existing GAN architecture known as styleGAN [30]. Unlike regular GAN, styleGAN uses a mapping network to map points in the latent space to an intermediatory latent space to control the style of the data in the generator. Furthermore, each point in the generator is provided with a noise vector for variation. This enables the model to produce higher quality images by allowing control over the style and offer greater diversity by using the noise vector. Figure 6 illustrates the some of the generated samples. The authors did not evaluate the performance of the generations but rather introduced the dataset and encouraged researchers to work on improving the quality of the generations.

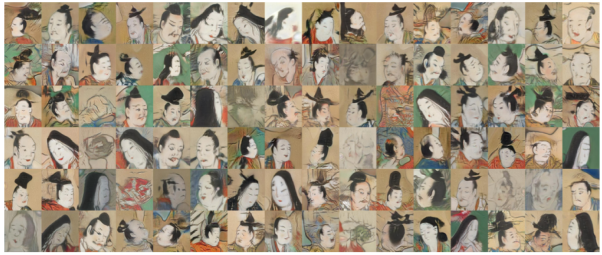

Figure 6 Samples from Japanese Art Facial Expression Generated using StyleGAN [29]

A framework for generating a face photo from a given sketch as well as a sketch from a given face photo is implemented using GANs in [31]. The training data used contained labeled face and sketch pairs from two different datasets which were combined. The authors trained a simple conditional GAN for both applications. Apart from presenting the generator and discriminator losses, no further evaluation was performed. However, comparison between face sketch produced by some of the previous works, a real artist, and the proposed work was presented.

Zheng *et al.* [32] introduced StrokeNET, a GAN-based architecture to generate digits and character strokes. The generator takes a single stroke, which contains the coordinate as well as pressure values, as an input. This is followed by a position encoder that encodes the coordinates and a brush encoder that encodes the radius of the brush. The feature maps produced by each encoder is then concatenated and passed on to a deconvolutional layer. The generator produces a 265 by 265 image output. Several datasets including MNIST was used to evaluate the network. The paper also presented the results of classification between the original MNIST digits, and the ones generated by the proposed model using a 5-layer CNN classifier. The reported classification accuracy using the pre-processed images was 91%. Besides handwriting, the network was also used to generate small sketches. A similar application of generating handwritten digits and calligraphy was presented in [33]. Two encoders were first used to encode the calligraphic style and the textual content to be written before feeding in the encoding to the generator. The discriminator simply evaluates between the ground truth and the generated sample. The best FID score obtained was 120.1 in comparison to the FID score of 90.4 for real images. Human evaluators were also asked to assess whether the generated images were sketched by humans or computer-generated. From 12k responses collected from 200 human examiners, only 49.3% of the images were correctly identified to be computer-generated. Moreover, the paper also presented a comparison of sample generation with one of the previous works, as shown in Figure 7. For a comprehensive survey of freehand sketches using deep learning, the readers are encouraged to refer to the following work [13].

Figure 7 Sample Handwriting Generation in [33] and Comparison with a Previous Work

This section focused on the recent applications of GAN with regards to the generation of visual arts. Table 1 provides a summary of the recent works in this domain along with their approaches and results. Furthermore, the table also summarizes the GAN type used, the loss function as well as the generator and discriminator architectures.

Table 1 Recent Advances in Visual Arts Generation using GANs

| Source | Task | GAN Type | Loss Function | Generator-Discriminator Architecture | Result |
|---|---|---|---|---|---|
| [24] | Generate cartoon image from sketch | Conditional GAN | Cross entropy, L1 distance for pixel-level loss | U-Net Generator, CNN-FC discriminator | Qualitative only, outperformed existing works |
| [25] | Generate shoe image from sketch | Conditional GAN | Binary cross-entropy loss for discriminator | U-Net Generator, Deep CNN discriminator | No evaluation |
| [26] | Generate fully colored synthetic images from sketch | Vanilla GAN with Encoder for image style recognition | Auxiliary Losses, discriminator is trained on style loss & content loss. MSE Loss. | U-Net Generator, DNN discriminator | FID 4.18, classification score 0.57. 63% evaluators preferred images from proposed model |
| [27] | Generate painting by brushstrokes | Conditional GAN | Wasserstein Loss | Configurations not provided | No evaluation |
| [28] | Unconditional art generation | DCGAN | Cross entropy loss and added classification, style ambiguity losses | Deep CNNs, CONV followed by LeakyReLU | 53% of evaluators believe that synthesized images were by an artist. |
| [29] | Generate pre-modern Japanese art facial expression | StyleGAN | WGAN-GP | Configurations not provided | No evaluation |
| [31] | Generate face photo from a sketch and vice versa | Conditional GAN | Cross entropy | Configurations not provided | No evaluation, the loss values were reported |
| [32] | Generate digits and character strokes | Modified DCGAN + agent | MSE | Generator contains CONV+LeakyReLU, Agent is VGG | Generated images not evaluated, classification accuracy 91% on MNIST |
| [33] | Generate calligraphy and handwritten digits | Modified conditional GAN with multiple encoders for style & content-encoding | Binary Cross entropy discriminator, cross-entropy style loss & Kullback-Leibler content loss | Two residual blocks then 4 CONV modules generator, 1 CONV layer then 6 residual blocks discriminator | FID 120.1, 49.3% of the images were identified to be synthesized |

## C. Music and Melody Generation using GANs

Next, GAN approaches for music and melody generation are presented. Welikala and Fernando [34] used Musical Instrument Digital Interface (MIDI) files, which contain data to specify the musical instruction such as note's notation, pitch, and vibrato, to train a hybrid variational autoencoder (VAE) and GAN to generate musical melody for a specific genre. In this work, five genres were considered, namely folk music, Arabic, jazz, metal rock, and classical. The total number of samples after adding all the genres was 1998 (10 seconds MIDI files) from the Nottingham dataset [35]. The encoder first encodes the previous frame to its corresponding latent representation, which is used by the generator for generating the next frame. This ensures that the generator receives the information from previous frames and is using it for the prediction of the next frame. The discriminator is then presented the real training frame as well as the synthesized one. Human evaluators consisting of expert musicians as well as amateurs were provided three samples to evaluate, and these samples were generated with three different epochs. However, no quantitative evaluation was presented. It was concluded that the consistency of generated melodies was not up to the same level as human composition. Similarly, [36] proposed a melody generation framework using GAN, consisting of a Bi-directional LSTM generator and an LSTM discriminator. Instead of MIDI files, they processed XML files. Furthermore, Bayesian optimization was selected to determine the hyperparameters of the next sample by considering the distribution using Bayes' theorem. This allowed the information to be extracted from previous sample points. Finally, nineteen participants were invited to evaluate the quality of the music. They were asked to rank each model from 1 to 5 based on how pleasing, realistic, and interesting the music sounded. The proposed model on average obtained a score 3.27 on the three metrics, outperforming the baseline models. Moreover, the samples generated by the proposed model was likely to be detected as synthesized by human evaluators only 48.1% of the time.

The authors in [37] presented an adversarial and convolutional based architecture known as MidiNet for generating pop music monophonic melodies using 1022 pop music from an online MIDI database called *TheoryTab* [38]. The dataset provides two channels for each tab, one for melody and the other one for the underlying chord progression. This allows training either with only the melody channel or using chord information to condition the generation. To generate additional training examples, augmentation was performed by circularly shifting all melodies and chords to any of the 12 keys, resulting in a final dataset of 50,496 melody and chord pairs. Both the generator and discriminator were convolutional networks. The generator consists of six layers: two dense layers followed by four convolutional layers. The generator, which generates the samples iteratively, is conditioned on the historical information from previous measures and the sequence of the chord. The discriminator includes two convolutional layers followed by fully connected layers. A case study with 21 participants, including 10 people with a music background, was conducted to evaluate the generations. The participants were asked to rate the generations from 1 to 5 based on how pleasing, real, and interesting they sounded. The mean value was around 3 for being pleasant and realistic. Moreover, the mean value in being interesting was around 4 for people with musical backgrounds, and 3.4 for people without musical backgrounds for the proposed MidiNet model. Yu *et al*. [39] used conditional GANs for melody generation. The model consists of three layers. The lyrics encoder tokenizes the lyrics into a vector containing a series of word and syllable tokens as well as additional noises. The output of the layer is fed into the conditioned-lyrics GAN. The generator uses LSTM to study the sequential alignment between the lyrics and melody and the real MIDI samples' distribution. The discriminator tries to discriminate between the real MIDI samples from the training set and the generated samples. MIDI sequence turner was then used to quantize the MIDI notes produced by the generator. The *Lakh MIDI* [40] and Reddit MIDI datasets were combined, which resulted in a total of 12,197 MIDI files. The authors created a GUI that produces a synthesized musical sheet using the piano MIDI number and human voice imitation. A similar study was conducted in [41] using the exact dataset and conditional-LSTM GANs. Comparison with random and maximum likelihood estimation baselines showed that the proposed model outperformed the baselines in terms of BLEU score. Qualitative evaluation showed that the melody score for the proposed model is 4.1 out of 5 and is very close to the ground truth melodies.

Mogren [42] proposed an RNN GAN for generating single voice polyphonic music. Using MIDI files, the objective was to model the note using the duration, tone, intensity, and time elapsed since the last tone. The dataset used consists of 3697 classical music from 160 different composers. The architecture of both the generator and discriminator contains 2 LSTM layers of 350 hidden units. In addition, the discriminator contains a bidirectional recurrent layout to take into account the context for the decision. Although the author has made the generated music available online, no evaluation with regards to the quality of the generated music was presented. A simple graphical analysis, in terms of statistics such as scale consistency and intensity, was provided along with comparison with the maximum likelihood estimation baseline. Dong *et al.* [43] introduced a novel multi-track polyphonic symbolic music generator using GANs. The proposed 'MuseGAN' architecture utilized three different GANs namely the jamming model, the composer model, and the hybrid model. The inspiration behind the jamming model was to model musicians improvising music without pre-defined arrangement. The inspiration behind the composer model was to arrange instruments with knowledge of harmonic structure. Meanwhile, the hybrid model combined the best of both. To implement the GANs, the Lakh MIDI dataset was transformed into a multi-track piano-rolls representation. The generators utilized 1D transposed convolutional layers whereas the discriminators contained five 1D convolutional layers followed by a dense layer. Various intra-track and inter-track evaluations were performed for each model including drum pattern and tonal distance. For intra-track metrics, the jamming model provided the best performance whereas for inter-track, the hybrid and composer performed better. A qualitative case study consisting of 144 participants provided an overall score of 2.86 out of 5 for the hybrid model.

This section focused on the recent applications of GAN with regards to music and melody generation. Table 2 provides a summary of the recent works in this domain along with their approaches and results. Moreover, the table also summarizes the GAN type used, the loss function as well as the generator and discriminator architectures.

Table 2 Recent Advances in Music and Melody Generation using GANs

| Source | Task | GAN Type | Loss Function | Generator-Discriminator Architecture | Result |
|---|---|---|---|---|---|
| [34] | Generate melody for a specific genre | Hybrid VAE and GAN | KL Divergence | Four DeCONV layers for both generator & discriminator, VAE encoder used three Conv2D | No quantitative evaluation, concluded that consistency of generated melodies was not up to the same level as human composition |
| [36] | Melody Generation | LSTM-based GAN | Bayesian | Bi-LSTM generator and LSTM discriminator | Average score of 3.27 on the three qualitative metrics, 48% likely to be detected as synthetic |
| [37] | Generate pop music monophonic melodies | Modified DCGAN | Cross entropy | Two dense layers followed by four transposed CONV for generator; 2 CONV layers followed by a dense layer discriminator | Mean score around 3 for being pleasant & realistic, 4 for interesting people with musical backgrounds, 3.4 for people without musical backgrounds |
| [39] | Generate melodies based on lyrics | Conditional GAN | Cross entropy | LSTM generators and discriminators | No evaluation was provided |
| [41] | Generate melodies based on lyrics | Conditional LSTM GAN | Cross entropy | Dense layer followed by 2 LSTM followed by a dense layer for generator, 2 LSTM followed by dense for discriminator | BLEU-2 score of 0.735, scores of about 3.8, 3.5, 4.1 respectively out of 5 for lyrics, rhythm, and melody by evaluators |
| [42] | Generate single voice polyphonic music | RNN GAN | Cross entropy and Squared error loss | 2 LSTM layers for generator, 2 Bi-directional LSTM layers followed by a dense for discriminator | No evaluation was provided |
| [43] | Generate multi-track, polyphonic music | Conditional GAN | Wasserstein | Generators contain 1D transposed CONV, discriminators 5 contain 1D Conv layers followed by one dense layer | The highest score for conditional generation was 3.1 and non-conditional was 3.16 out of 5 by 'non-pro' evaluators. For intra-track metrics, jamming model performed best |
| [44] | Generate folk music | RL GAN | Cross entropy and policy gradient | RNN generators, CNN discriminators | BLEU score of 0.94 and MSE of 20.6 outperformed baseline maximum likelihood estimation |

## D. Poetry and Literary Text Generation using GANs

Researchers have also focused on literary text generation using GANs. This section presents a comparison of GANs in poetry and literary text generation. Most GAN architectures are restricted by several factors when it comes to text and sequential generation. For instance, the feedback given by the discriminator applies to the entire sequence and the generations are usually continuous data. Yu *et al.* [44] introduced SeqGAN to overcome these challenges by defining the generator as stochastic policy in reinforcement learning (RL). Moreover, the reward in RL arrives from the discriminator which is evaluated on the entire sequence. Then, it is passed back to the intermediate state-action pairs using the Monte Carlo search algorithm. The SeqGAN architecture is illustrated in Figure 8.

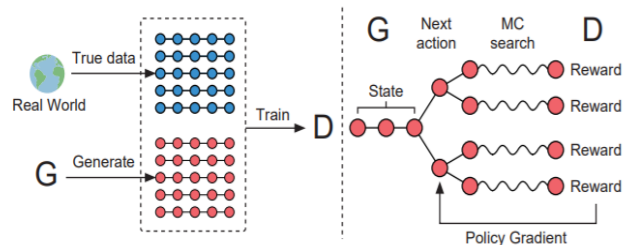

Figure 8 SeqGAN Architecture [44]

The discriminator in SeqGAN is trained using both the real as well as generated data. More specifically, the positive class examples are taken from the real data and the negative class are from the data synthesized by the generator. The generator is simultaneously trained by RL approach and Monte Carlo search based on the expected reward obtained from the discriminator, which is the likelihood that it would fool the discriminator. Among several experiments using this architecture, the authors also employed the model for Chinese poem generation. The dataset used comprises of over 16, 000 poems with each poem containing four lines of twenty characters. BLEU score was used to measure the similarity of the synthesized text and the human-created ones. The SeqGAN generated poems outperformed the baseline maximum likelihood estimation on the BLEU score (statistically significant with p-value less than $10^{-6}$). Additionally, 70 experts on Chinese poems were asked to evaluate between 20 real poems, 20 generated using maximum likelihood estimation, and 20 generated using SeqGAN. The SeqGAN outperformed the baseline with a 0.54 average score and a statistically significant p-value. The approach of extending GANs to generate sequences of discrete tokens is suitable for poetry and other text generations because of the sequential nature of text datasets.

An end-to-end framework for generating poetry from an image is presented in [45]. As part of this work, the dataset collected for pairing image and poem by human annotators is made available online. The generative model is a CNN-RNN architecture, acting as an agent. The primary parameters of the agent describe the policy to decide the words to be selected for a given action. The reward is given when the agent is able to pick all the words in a poem. The first of the two discriminative models provides rewards based on whether the generated poem is correctly paired with the input image whereas the second model provides rewards based on whether the synthesized poem is 'poetic'. The proposed image to poetry GAN (I2P-GAN) was evaluated on several metrics such as relevance, novelty, and BLEU scores against different architectures including SeqGAN. Overall, it outperformed all the models with an average score of 77.2% which is the normalized score considering the average of all the other scores. Human evaluators were also invited to evaluate out of 10 on relevance, coherence, imaginativeness, and the generations overall. The proposed I2P-GAN outperformed all the models with a 7.18 overall score which is close to the ground truth overall score of 7.37. Figure 9 shows a comparison of a sample image and generations from the proposed and baseline models.

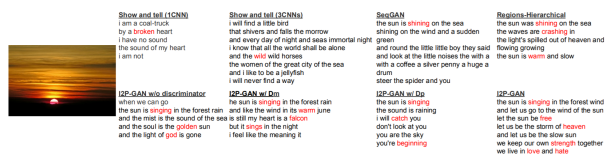

Figure 9 Poems Generated from a Given Image of the Proposed I2P-GAN in [45] and Comparison with Baseline Models

Kashyap *et al.* [46] proposed a Shakespearean prose generation from a given painting. Due to the dataset of painting to Shakespearean prose not being available, the authors utilized intermediate English poem description of the paintings before applying language style transfer to obtain the relevant prose style. A total of three datasets were used to solve this application. Two of the datasets were used for generating an English poem for a given image. For text style transfer, Shakespeare plays, and their corresponding English translation dataset was used. The architecture used is displayed in Figure 10.

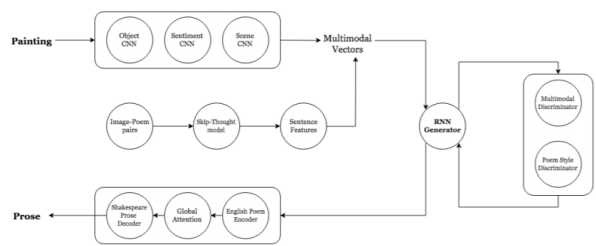

Figure 10 Proposed Prose Generation Architecture in [46]

The poem generation utilizes three parallel CNNs for extracting the object, sentiment, and scene features, respectively. To obtain poetic indications, the features are then merged with a skip-thought model. This is followed by a sequence-to-sequence model trained by policy gradient and two discriminators providing feedback rewards. This architecture of the CNN-RNN model as an agent and two discriminators is similar to that of [45]. The overall goal of the poem generation architecture is to generate a sequence of words as a poem for an image by maximizing the expected return. To obtain the final prose, a sequence-to-sequence model with a unidirectional single layer LSTM encoder and a single layer LSTM decoder with pre-trained word embeddings were used. For evaluation, 32 students were asked to score the generations from 1-5 based on content, creativity, and similarity to Shakespearean style. The average scores for each parameter were 3.7, 3.9, and 3.9, respectively. Experiments also showed that as source sentence length increases, the BLEU score decreases. Another limitation of this approach is that the generations may suffer in quality when the style transfer dataset does not have similar words in the training set of sequences and consequently, the dataset must be expanded. A sample generation of the poem (center) and prose (right) of the proposed model is shown in Figure 11.

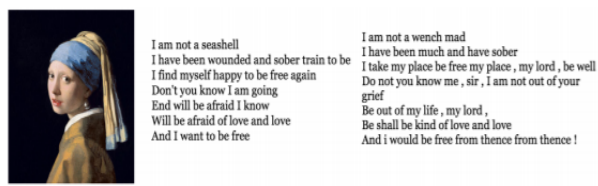

Figure 11 A sample Poem and Prose Generation from [46]

GANs are excellent at modeling continuous distributions, making them suitable for tasks like image generation. However, their application towards language modeling which are in discrete settings is limited as a result of complexity in backpropagation through discrete random variables. As a result, a maximum likelihood augmented discrete GAN was proposed in [47]. This architecture introduced a low-variance objective function using the discriminator's result following the corresponding log-likelihood. The proposed model was experimented on several tasks including poetry generation. For poetry generation, two Chinese poem datasets, named Poem-5 and Poem-7, were used with each containing 5 or 7 characters in short sentences, respectively. The generator is a single layer LSTM and discriminators are two-layered Bi-directional LSTMs. The proposed model outperformed several baselines such as SeqGAN in terms of several metrics including BLUE score. However, the authors did not provide a qualitative evaluation. Moreover, no samples of the generated poems were presented. Similarly, [48] handled the discrete output space issue by forcing the discriminator to operate on continuous-valued output distributions. A sequence of probabilities for every token in the vocabulary from the generator as well as a sequence of 1-hot vectors from the real data distribution was presented to the discriminator. For objective function, Wasserstein GAN (WGAN), as well as WGAN with gradient penalty (WGAN-GP) [49], were considered. Both LSTM-based, as well as CNN-based GAN architectures, were experimented and the proposed model with LSTM outperformed the existing works on the aforementioned Poem-5 and Poem-7 datasets. Despite presenting the BLEU score evaluations, no qualitative case study or results samples were presented in this work.

Lin *et al.* [50] presented RankGAN by taking inspiration from the learning to rank concept from information retrieval, where given a reference, the required information is integrated into the ranking function encouraging relevant documents to be returned quicker. The overall architecture of RankGAN is presented in Figure 12.

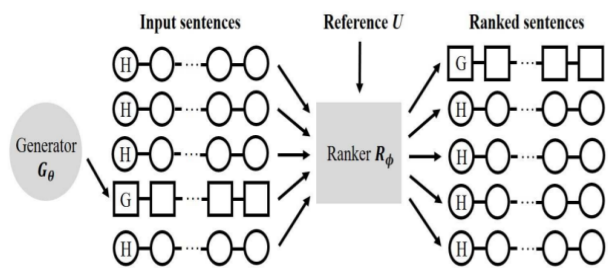

Figure 12 Proposed RankGAN Architecture in [50]

The RankGAN differs from traditional GAN by including a sequence of generators and a ranker. The ranker is responsible for providing relative ranking among sequences, given a reference. The objective function of the generator is to generate a sentence that can potentially obtain a greater ranking score than the ones taken from the real dataset. On the other hand, the ranker's task is to rank the generated sentence to be of lower importance than the sentences from the real dataset. The proposed architecture was evaluated for poetry generation on a Chinese poem dataset containing over 13,000 five-word quatrain poems. In terms of BLUE score, the proposed RankGAN outperformed both SeqGAN and maximum likelihood estimation. Additionally, 57 native mandarin Chinese speakers were asked to rate the generations from SeqGAN, RankGAN, and human-written poems. The average scores provided by the evaluators were 3.4, 4.6, and 6.4 for SeqGAN, RankGAN, and human-written poems, respectively. The proposed RankGAN model was also applied to Shakespeare's Romeo and Juliet play to learn lexical dependency and use rare phrases in the play. The high BLEU-2 score of 0.91 indicates that RankGAN was able to capture the transition pattern among the words despite the training sentences being novel, subtle, and complex. No qualitative evaluation or generated samples were presented for this application.

Saeed *et al.* [51] proposed a GAN architecture for creative text generation. Instead of optimizing maximum likelihood estimation, which produces generic and repetitive generations, the goal in this research was to generate diverse and unique samples. This was achieved using a discriminator to provide feedback to the generator by a cost function that encourages sampling of creative tokens. The generator in this framework is a language model trained using backpropagation through time. A pre-training phase is included where maximum likelihood estimation is optimized. Then, during the training phase, the creativity reward from the discriminator is optimized. The discriminator contains an encoder-decoder pair with the encoder having the same architecture as the generator, with an added pooled decoder layer. The decoder is composed of three blocks of dense batch normalization and ReLU activation, followed by a sigmoid layer. The dataset used for poetry generation consists of 740 classical and contemporary English poems. In terms of perplexity scores, the proposed Creative-GAN outperformed the existing works. However, the authors did not provide a qualitative analysis of the results as well as sample generation of texts.

This section presented a comparison of recent applications of GAN with regards to poetry and literary text generations. Table 3 provides a summary of the recent works in this domain along with their approaches, and results. Moreover, the GAN type, loss function, and architectural details are also highlighted.

Table 3 Recent Advances in Literary Text Generation using GANs

| Source | Task | GAN Type | Loss Function | Generator-Discriminator Architecture | Dataset | Result |
|---|---|---|---|---|---|---|
| [44] | Chinese Poetry generation | RL GAN | Cross entropy and policy gradient | RNN Generators and CNN discriminators | 16394 Chinese quatrains | BLEU-2 score of 0.74, overall score of 0.54 by human evaluators |
| [45] | Generate poetry from an image | Multiadversarial GAN with an embedding model | Cross entropy and policy gradient | RNN Generators, GRU-based discriminator, CNN image encoder and RNN poem decoder | Novel dataset with paired image and poetry | Overall BLEU score of 0.77, 7.18 out of 10 overall score by human evaluators |
| [46] | Generate Shakespearean prose from a painting | Multiadversarial GAN with encoder and decoder | Cross entropy and policy gradient | RNN Generator, CNN-RNN agent for encoding and decoding painting, LSTM encoder and decoder for generating prose | Two datasets for generating English poem from an image, and Shakespeare plays and their English translations for text style transfer | Average scores of 3.7, 3.9, and 3.9 out of 5 by evaluators for content, creativity, and similarity to Shakespearean style respectively |
| [47] | Chinese Poetry generation | RL GAN | Maximum-likelihood | A single layer LSTM generator, two-layer Bi-directional LSTMs discriminator | Poem-5 and Poem-7 Chinese Poem dataset | BLEU-2 scores of 0.76 and 0.55 for the two datasets respectively |
| [48] | Chinese Poetry generation | RNN GAN | Wasserstein distance | LSTM generator and discriminator with WGAN-GP training | Poem-5 and Poem-7 Chinese Poem dataset | BLEU-2 scores of 0.88 and 0.67 for the two datasets respectively |
| [50] | Chinese Poetry generation | GAN with a ranking function | Ranking objective | LSTM generator, CNN-based ranker | Over 13,000 Chinese quatrains | BLEU-2 score of 0.81, 4.6 out of 10 overall score by human evaluators |
| [50] | Learn rare words from Romeo and Juliet play | GAN with a ranking function | Ranking objective | LSTM generator, CNN-based ranker | Over 3000 sentences from Romeo and Juliet play | BLEU-2 score of 0.914 |
| [51] | Poetry and lyrics generation | GAN with language model generator | Cross entropy | AWD-LSTM [52] and TransformerXL [53] language model for generator, discriminator encoder-decoder pair. | 740 classical and contemporary English poems and 1500 song lyrics across various genres | Perplexity score of 42.5 for poetry and 9.02 for lyrics generations |

## IV. CHALLENGES AND FUTURE WORK RECOMMENDATIONS

Although the previous section provided some promising recent developments in generating arts using GANs, there remains a few challenges. In this section, we look at the major challenges for each art form and highlight recommendations for future work.

### A. Challenges

For visual arts generation, most notably, there does not exist a consistent approach for performing qualitative validation among the existing works. Although generally, the qualitative surveys are being conducted to find out if the generated arts are perceived to be real or synthesized, there needs to be a consistent qualitative assessment on the creative aspect of the generated arts. Research works on visual art generations should be encouraged to invite artists to provide an assessment of the creativity. Moreover, the size of the dataset required for GAN training remains a challenge. The lowest dataset size from the existing works reviewed in Table 1 utilized at least 5000 images [29]. Therefore, for novel applications, a lot of time is required in gathering the dataset first before applying the GAN training. This can therefore be a huge challenge in terms of time and resource constraints needed for the implementation of such projects.

There remain a few notable challenges in training GANs for music and melody generation. Firstly, music is time-dependent by nature. For instance, the set of musical notes to be played depends on the previous set of notes. Therefore, unlike generating images, music generation requires a more sophisticated approach that can consider the sequential nature of the data. Moreover, as summarized in Table 2, most of the existing works dealt with MIDI file format to generate melodies. This makes it challenging for non-specialists in music to work with music generation applications. This is different from image generation tasks where there is a convention of directly using the image files for model training. Finally, there are no standard evaluation approaches for assessing the quality of the generated melodies. Rather, most of the existing works have preferred qualitative assessments using human evaluators. This makes it challenging in providing a comparative study of the recent works.

There are two major challenges for literary text generation. Firstly, most of the reviewed works in Table 3 addressed poetry generation application. Only a handful of prose and lyrics generation applications can be found in existing works. Therefore, to truly determine the ability of GAN with written arts, other applications such as novel writing and rhyme generations should be explored. Secondly, there is a lack of consistent methodology for qualitative evaluation as a lot of the existing works did not even perform a qualitative evaluation. In terms of quantitative measure, BLUE score and Perplexity scores are widely accepted for evaluating text generations. Therefore, researchers should be encouraged to invite poets and writers to evaluate the generated literary texts on their creativity.

### B. Recommendations and Future Work

In the context of visual arts generation, an experiment with smaller datasets should be carried out. This would lead to a better understanding of how well GANs can work for generating visual arts when the data is scarce. It could potentially lead to the development of a GAN framework that is well suited to dealing with such datasets. This is important because artworks, especially of the pre-modern times are rare. Therefore, training large scale GAN architectures will not be suitable. Furthermore, a standard qualitative assessment approach should be proposed that can be effectively used by all visual arts generating applications. This should include guidelines in incorporating both specialists and non-specialists in arts for providing their feedback. An adequate sample size that would lead to statistically significant results should also be considered for qualitative evaluation.

For music and melody generation applications, most of the previous works have focused on using the MIDI file format for generating music. However, this makes it challenging for non-specialists in music to get involved and improve upon the music generation. Therefore, future research should focus on implementing GAN architectures for generating music in raw audio format. The major challenge for this approach will be the computation resources in dealing with raw audio files. Furthermore, there does not exist a well-defined metric in terms of quantitative evaluation that can be used to assess the quality of the generated music. Therefore, it is recommended to carry out a study that can define new metrics which measures the quality of music generation adequately.

In terms of poetry and other literary text generations, a great emphasis on poetry and small text generations can be observed. Therefore, to find out the effectiveness of GANs in producing larger text pieces such as novels and dramas, researchers are encouraged to take on the challenge with longer texts. Moreover, despite there being an agreed convention of using BLEU and Perplexity scores for quantitative evaluation, a well-defined qualitative evaluation method should be proposed for meaningful comparison. This would allow for a more comprehensive assessment of the generated text pieces.

The previous section of this paper clearly highlighted the remarkable progress in recent years for generating various artworks utilizing GANs. This demonstrates great opportunities and potential for computer art generation research. In summary, for future work on art generation using GANs, we recommend the following:

- Experiment with smaller dataset and GAN architectures for visual arts generation.
- Introduce a standard qualitative validation approach for visual arts generation.
- Implement GANs to generate music in raw audio format as opposed to MIDI file format.
- Introduce a quantitative measure for evaluating the quality of the music generated.
- Implement longer text literary work generations including novels and dramas.
- Propose a well-defined and comprehensive qualitative validation approach for literary text generation.

## V. CONCLUSION

In this survey, a historical background on computer generated arts was first presented along with the limitations. Fortunately, due to the advancement in the field of deep learning, GANs emerged as a promising technology for computer generated arts. After a brief overview on different types of GANs and loss functions, the paper presented the recent works on generating visual arts, music, and literary texts using GANs. A discussion on the results as well as challenges and future research direction was also provided.

REFERENCES

[1] G. Vasari, *The lives of the most excellent painters, sculptors, and architects*. Modern Library, 2007.

[2] S. Davies, "Definitions of art," *The Routledge companion to aesthetics*, pp. 169–179, 2001.

[3] M. A. Boden and E. A. Edmonds, "What is generative art?," *null*, vol. 20, no. 1–2, pp. 21–46, Jun. 2009, doi: 10.1080/14626260902867915.

[4] S. LeWitt, "Paragraphs on conceptual art," *Artforum*, vol. 5, no. 10, pp. 79–83, 1967.

[5] J. Vincent, "A look back at the first computer art contests from the '60s: bullet ricochets and sine curve portraits," *The Verge*, Jul. 13, 2015. https://www.theverge.com/2015/7/13/8919677/early-computer-art-computers-and-automation (accessed Apr. 14, 2021).

[6] E. C. Berkeley, "Computer Art Contest," *Computers and Automation*, 1968.

[7] A. M. Noll, "The digital computer as a creative medium," *IEEE spectrum*, vol. 4, no. 10, pp. 89–95, 1967.

[8] S. Shahriar and U. Tariq, "Classifying Maqams of Qur'anic Recitations using Deep Learning," *IEEE Access*, pp. 1–1, 2021, doi: 10.1109/ACCESS.2021.3098415.

[9] I. Goodfellow, Y. Bengio, and A. Courville, *Deep learning*. MIT press, 2016.

[10] J. Brownlee, "A gentle introduction to generative adversarial networks (GANs)," *Retrieved June*, vol. 17, p. 2019, 2019.

[11] S. DiPaola and L. Gabora, "Incorporating characteristics of human creativity into an evolutionary art algorithm," *Genetic Programming and Evolvable Machines*, vol. 10, no. 2, pp. 97–110, 2009.

[12] I. Santos, L. Castro, N. Rodriguez-Fernandez, Á. Torrente-Patiño, and A. Carballal, "Artificial Neural Networks and Deep Learning in the Visual Arts: a review," *Neural Computing and Applications*, vol. 33, no. 1, pp. 121–157, Jan. 2021, doi: 10.1007/s00521-020-05565-4.

[13] P. Xu, "Deep learning for free-hand sketch: A survey," *arXiv preprint arXiv:2001.02600*, 2020.

[14] I. Goodfellow *et al.*, "Generative adversarial nets," *Advances in neural information processing systems*, vol. 27, 2014.

[15] M. Mirza and S. Osindero, "Conditional generative adversarial nets," *arXiv preprint arXiv:1411.1784*, 2014.

[16] A. Radford, L. Metz, and S. Chintala, "Unsupervised representation learning with deep convolutional generative adversarial networks," *arXiv preprint arXiv:1511.06434*, 2015.

[17] A. Creswell, T. White, V. Dumoulin, K. Arulkumaran, B. Sengupta, and A. A. Bharath, "Generative adversarial networks: An overview," *IEEE Signal Processing Magazine*, vol. 35, no. 1, pp. 53–65, 2018.

[18] D. J. Im, C. D. Kim, H. Jiang, and R. Memisevic, "Generating images with recurrent adversarial networks," *arXiv preprint arXiv:1602.05110*, 2016.

[19] X. Chen, Y. Duan, R. Houthooft, J. Schulman, I. Sutskever, and P. Abbeel, "Infogan: Interpretable representation learning by information maximizing generative adversarial nets," in *Proceedings of the 30th International Conference on Neural Information Processing Systems*, 2016, pp. 2180–2188.

[20] E. Denton, S. Chintala, A. Szlam, and R. Fergus, "Deep generative image models using a laplacian pyramid of adversarial networks," *arXiv preprint arXiv:1506.05751*, 2015.

[21] M. Arjovsky, S. Chintala, and L. Bottou, "Wasserstein generative adversarial networks," in *International conference on machine learning*, 2017, pp. 214–223.

[22] A. Bissoto, E. Valle, and S. Avila, "The six fronts of the generative adversarial networks," *arXiv preprint arXiv:1910.13076*, 2019.

[23] "Wasserstein metric," *Wikipedia*. Apr. 23, 2021. Accessed: May 26, 2021. [Online]. Available: https://en.wikipedia.org/w/index.php?title=Wasserstein_metric&oldid=1019498722

[24] Y. Liu, Z. Qin, Z. Luo, and H. Wang, "Auto-painter: Cartoon image generation from sketch by using conditional generative adversarial networks," *arXiv preprint arXiv:1705.01908*, 2017.

[25] B. Kuriakose, T. Thomas, N. E. Thomas, S. J. Varghese, and V. A. Kumar, "Synthesizing Images from Hand-Drawn Sketches using Conditional Generative Adversarial Networks," in *2020 International Conference on Electronics and Sustainable Communication Systems (ICESC)*, 2020, pp. 774–778. doi: 10.1109/ICESC48915.2020.9155550.

[26] B. Liu, K. Song, Y. Zhu, and A. Elgammal, "Sketch-to-Art: Synthesizing Stylized Art Images From Sketches," 2020.

[27] R. Nakano, "Neural painters: A learned differentiable constraint for generating brushstroke paintings," *arXiv preprint arXiv:1904.08410*, 2019.

[28] A. Elgammal, B. Liu, M. Elhoseiny, and M. Mazzone, "Can: Creative adversarial networks, generating" art" by learning about styles and deviating from style norms," *arXiv preprint arXiv:1706.07068*, 2017.

[29] Y. Tian, C. Suzuki, T. Clanuwat, M. Bober-Irizar, A. Lamb, and A. Kitamoto, "KaoKore: A Pre-modern Japanese Art Facial Expression Dataset," *arXiv preprint arXiv:2002.08595*, 2020.

[30] T. Karras, S. Laine, and T. Aila, "A style-based generator architecture for generative adversarial networks," in *Proceedings of the IEEE/CVF Conference on Computer Vision and Pattern Recognition*, 2019, pp. 4401–4410.

[31] C. Philip and L. H. Jong, "Face sketch synthesis using conditional adversarial networks," in *2017 International Conference on Information and Communication Technology Convergence (ICTC)*, 2017, pp. 373–378. doi: 10.1109/ICTC.2017.8191006.

[32] N. Zheng, Y. Jiang, and D. Huang, "Strokenet: A neural painting environment," 2018.

[33] L. Kang, P. Riba, Y. Wang, M. Rusiñol, A. Fornés, and M. Villegas, "GANwriting: Content-Conditioned Generation of Styled Handwritten Word Images," in *Computer Vision – ECCV 2020*, Cham, 2020, pp. 273–289.

[34] M. D. Welikala and T. Fernando, "Komposer V2: A Hybrid Approach to Intelligent Musical Composition

Based on Generative Adversarial Networks with a Variational Autoencoder," in *Proceedings of the Future Technologies Conference*, 2020, pp. 413–425.
[35] "jukedeck/nottingham-dataset," *GitHub*. https://github.com/jukedeck/nottingham-dataset (accessed May 25, 2021).
[36] Y. Xu, X. Yang, Y. Gan, W. Zhou, H. Cheng, and X. He, "A Music Generation Model Based on Generative Adversarial Networks with Bayesian Optimization," in *Chinese Intelligent Systems Conference*, 2020, pp. 155–164.
[37] L.-C. Yang, S.-Y. Chou, and Y.-H. Yang, "MidiNet: A convolutional generative adversarial network for symbolic-domain music generation," *arXiv preprint arXiv:1703.10847*, 2017.
[38] "Tabs that show the theory behind songs - Hooktheory." https://www.hooktheory.com/theorytab (accessed May 25, 2021).
[39] Y. Yu, F. Harscoët, S. Canales, G. Reddy, S. Tang, and J. Jiang, "Lyrics-conditioned neural melody generation," in *International Conference on Multimedia Modeling*, 2020, pp. 709–714.
[40] "The Lakh MIDI Dataset v0.1." https://colinraffel.com/projects/lmd/ (accessed May 25, 2021).
[41] Y. Yu, A. Srivastava, and S. Canales, "Conditional lstm-gan for melody generation from lyrics," *arXiv preprint arXiv:1908.05551*, 2019.
[42] O. Mogren, "C-RNN-GAN: Continuous recurrent neural networks with adversarial training," *arXiv preprint arXiv:1611.09904*, 2016.
[43] H.-W. Dong, W.-Y. Hsiao, L.-C. Yang, and Y.-H. Yang, "Musegan: Multi-track sequential generative adversarial networks for symbolic music generation and accompaniment," in *Proceedings of the AAAI Conference on Artificial Intelligence*, 2018, vol. 32, no. 1.
[44] L. Yu, W. Zhang, J. Wang, and Y. Yu, "Seqgan: Sequence generative adversarial nets with policy gradient," in *Proceedings of the AAAI conference on artificial intelligence*, 2017, vol. 31, no. 1.
[45] B. Liu, J. Fu, M. P. Kato, and M. Yoshikawa, "Beyond narrative description: Generating poetry from images by multi-adversarial training," in *Proceedings of the 26th ACM international conference on Multimedia*, 2018, pp. 783–791.
[46] P. Kashyap, S. Phatale, and I. Drori, "Prose for a Painting," *arXiv preprint arXiv:1910.03634*, 2019.
[47] T. Che *et al.*, "Maximum-likelihood augmented discrete generative adversarial networks," *arXiv preprint arXiv:1702.07983*, 2017.
[48] S. Rajeswar, S. Subramanian, F. Dutil, C. Pal, and A. Courville, "Adversarial generation of natural language," *arXiv preprint arXiv:1705.10929*, 2017.
[49] I. Gulrajani, F. Ahmed, M. Arjovsky, V. Dumoulin, and A. Courville, "Improved training of wasserstein gans," *arXiv preprint arXiv:1704.00028*, 2017.
[50] K. Lin, D. Li, X. He, Z. Zhang, and M.-T. Sun, "Adversarial ranking for language generation," *arXiv preprint arXiv:1705.11001*, 2017.
[51] A. Saeed, S. Ilić, and E. Zangerle, "Creative GANs for generating poems, lyrics, and metaphors," *arXiv preprint arXiv:1909.09534*, 2019.
[52] S. Merity, N. S. Keskar, and R. Socher, "Regularizing and optimizing LSTM language models," *arXiv preprint arXiv:1708.02182*, 2017.
[53] Z. Dai, Z. Yang, Y. Yang, J. Carbonell, Q. V. Le, and R. Salakhutdinov, "Transformer-xl: Attentive language models beyond a fixed-length context," *arXiv preprint arXiv:1901.02860*, 2019.